# An Energy Minimization Approach to 3D Non-Rigid Deformable Surface Estimation Using RGBD Data

Bryan Willimon, Steven Hickson, Ian Walker, and Stan Birchfield
Department of Electrical and Computer Engineering
Clemson University, Clemson, SC 29634
{rwillim,shickso,iwalker,stb}@clemson.edu

*Abstract*— We propose an algorithm that uses energy minimization to estimate the current configuration of a non-rigid object. Our approach utilizes an RGBD image to calculate corresponding SURF features, depth, and boundary information. We do not use predetermined features, thus enabling our system to operate on unmodified objects. Our approach relies on a 3D nonlinear energy minimization framework to solve for the configuration using a semi-implicit scheme. Results show various scenarios of dynamic posters and shirts in different configurations to illustrate the performance of the method. In particular, we show that our method is able to estimate the configuration of a textureless nonrigid object with no correspondences available.

## I. INTRODUCTION

Having concentrated on rigid objects for decades, roboticists have begun to turn their attention toward the manipulation of non-rigid objects. Among the several application areas motivating such investigation is that of automatic laundry handling. The ability of robots to locate clothing, identify clothing, manipulate and fold clothing, and so forth would be a tremendous asset to the increasingly promising field of domestic robotics. Notable progress has been made in recent years in developing systems capable of folding a T-shirt [1], [2], matching socks [3], and classifying articles of clothing [4], [5], [6], [7], [8], [9], [10], [11], among other tasks.

One of the core competencies in handling non-rigid objects such as clothing is the ability to estimate the configuration of the item, that is, a parameterization of the 3D coordinates of the various points on its surface. Unlike a rigid object, whose 3D pose is characterized by six values, a non-rigid item such as clothing has essentially infinite degrees of freedom, making its configuration much more difficult to recover. Nevertheless, the ability to estimate this configuration would aid in grasping and planning tasks as well as folding and unfolding.

In this paper we address the problem of estimating the 3D configuration of a known non-rigid object such as an article of clothing, piece of paper, or bendable poster. The item is modeled as a triangular mesh whose 3D coordinates completely specify the configuration of the item. Our approach formulates the problem as one of energy minimization, building on the work of Pilet et al. [12] by extending mesh estimation to 3D. By combining internal energy constraints with data terms that take into account the information available from an off-the-shelf RGBD sensor, the system is able to estimate the item's configuration in an efficient manner. The approach does not rely on modifying the item in any way (e.g., with handcrafted fiducial markers) but instead uses SURF feature correspondences along with novel depth and boundary terms to ensure that the mesh remains accurate even in untextured areas. Results on several different examples demonstrate the effectiveness of the approach.

## II. PREVIOUS WORK

Emerging research on non-rigid objects includes motion planning algorithms for deformable linear objects (DLOs) like ropes, cables, and sutures [13], [14]; Probabilistic RoadMap (PRM) planners for a flexible surface patch [15] or deformable object [16]; learning approaches to sense and model deformable surfaces [17], [18], [19]; and fabric manipulation for textile applications [20], [21], [22]. In particular, the problem of automating laundry has been receiving attention recently because of the increasing availability of calibrated two-handed robots like the PR2. Researchers have demonstrated systems for grasping clothes [4], [5], folding clothes [6], [1], [2], [3], [7], and tracing edges [8], [9]. Related research has addressed folding origami [23], folding towels [9], unfolding laundry [10], classifying clothing [24], [25], and pairing socks [11]. Despite the progress in manipulating non-rigid objects, none of this research addresses the problem of estimating the 3D geometry of a non-rigid object.

Two recent research projects have independently aimed at the goal of non-rigid 3D geometry estimation. Bersch et al. [26] use fiducial markers printed on a T-shirt. An augmented reality toolkit is used to detect markers, and stereo imagery is used to automatically generate a 3D triangular mesh. When the article of clothing is picked up, the grasp point is inferred from the distances from the end effector to the visible markers which then leads to a succeeding grasp point in a two-handed robotic system. Similarly, Elbrechter et al. [27] print specially designed fiducial markers on both sides of a piece of paper. Multi-view stereo vision is used to estimate the 3D positions of the markers which are then sent to a physics-based modeling engine. In addition to using two computers to control the two hands, three computers are used for data acquisition and processing.

A large amount of research has been conducted in the computer vision community in recent years on the problem of non-rigid structure from motion (NRSfM) for either inanimate objects [28], [29], people [30], [31], [32], or both [33], [34], [35]. These approaches utilize the movement of features within a 2D video sequence to recover the 3D coordinates of the preimages of the features in the scene. An alternate approach is to estimate a triangle "mesh soup" [35] or locally rigid patches [28] to yield a reconstruction in the form of a triangulation. In contrast with NRSfM, our goal is to register the 3D non-rigid model with the incoming RGBD sequence, similar to the 2D-3D registration of Del Bue and colleagues [36] [37].

In an effort to develop a system capable of handling real, unmodified objects, our work aims to remove the need for fiducial markers. Our approach is inspired by, and based upon, the research of Pilet et al. [12] which formulates the problem of 2D mesh estimation as energy minimization. In related work, Salzmann et al. [38], [39] describe an approach that operates in 3D but requires the restricted assumption of rigid triangles in order to reduce the dimensionality. Salzmann et al. [40] present a method for learning local deformation models for textured and textureless objects. Our approach extends this research by finding locally deformable 3D triangular meshes without fiducial markers whether intensity edges are present around the boundary of the object or not.

## III. APPROACH

Let $V = (v_1, \ldots, v_n)$ be the $n$ vertices of a 3D triangulated mesh, where $v_i = (x_i, y_i, z_i)$ contains the 3D coordinates of the $i$th vertex. The state vector $V$ captures the shape of the mesh at any given time, where we have omitted the time index to simplify the notation.

Our goal is to find the most likely mesh $V^*$ in each frame, by finding the shape that minimizes the energy of the system:

$$V^* = \arg\min \Psi(V). \tag{1}$$

We define the energy functional as the sum of four terms:

$$\Psi(V) = \Psi_S(V) + \lambda_C \Psi_C(V) + \lambda_D \Psi_D(V) + \lambda_B \Psi_B(V), \tag{2}$$

where $\Psi_S(V)$ is a smoothness term that captures the internal energy of the mesh, $\Psi_C(V)$ is a data term that seeks to ensure that corresponding points are located near each other, $\Psi_D(V)$ measures the difference in depth between the mesh vertices and the input, and $\Psi_B(V)$ is the boundary term that regulates the mesh vertices located on the boundary of the object. The weighting parameters $\lambda_C$, $\lambda_D$, and $\lambda_B$ govern the relative importance of the terms.

Each of the energy terms are useful in different scenarios. Figure 1 illustrates the contribution of each energy term for a toy example of a horizontal slice of vertices from the mesh. We now describe the energy terms in detail.

### A. Smoothness term

Let $\hat{V} = (\hat{v}_1, \ldots, \hat{v}_n)$, where $\hat{v}_i = (\hat{x}_i, \hat{y}_i, \hat{z}_i)$, be a hexagonal grid of equilateral triangles that is created either off-line or from the first image of the video sequence. We will refer to $\hat{V}$ as the *canonical* mesh. Except for the boundaries, each vertex in the canonical mesh has, when projected orthogonally onto the $z = 0$ plane, three pairs of collinear adjacent points passing through it, similar to

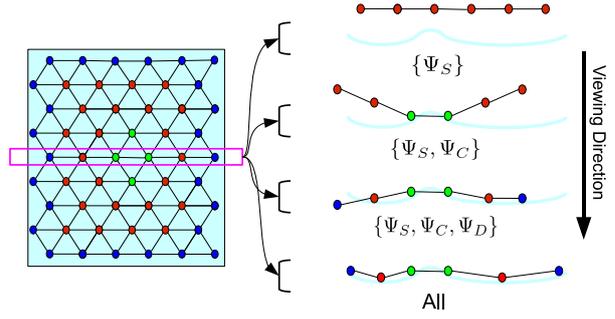

Fig. 1. Each of the energy terms contributes to improving the fit between the mesh and the object. Left: Front view of an example mesh, containing regular vertices (red dots), boundary vertices (blue dots), and vertices near image texture (green dots). Right: Top view of mesh. From top to bottom, the smoothness term ensures low derivatives in the mesh, the data term pulls the mesh toward the correspondences (green dots), the depth term moves vertices along the sensor viewing direction, and the boundary term introduces lateral forces to increase the fit at the boundaries (blue dots).

[12]. Let $E$ be the set of all vertex index triplets such that $(i, j, k) \in E$ means that $\hat{v}_i$, $\hat{v}_j$, and $\hat{v}_k$ form two connected, equidistant, and collinear edges in the projected canonical mesh. Since the projected canonical mesh is formed from equilateral triangles, we have

$$\hat{v}_i - \hat{v}_j = \hat{v}_j - \hat{v}_k \qquad \forall (i, j, k) \in E, \tag{3}$$

assuming that the initial configuration is approximately fronto-parallel. Therefore, in the deformed mesh $V$, we want this to approximately hold:

$$v_i - 2v_j + v_k \approx 0, \qquad \forall (i, j, k) \in E \tag{4}$$

which leads to the following smoothness term:

$$\Psi_S(V) = \frac{1}{2} \sum_{(i,j,k) \in E} (x_i - 2x_j + x_k)^2 \\ + (y_i - 2y_j + y_k)^2 \\ + (z_i - 2z_j + z_k)^2. \tag{5}$$

Let

$$X = \begin{bmatrix} x_1 & x_2 & \cdots & x_n \end{bmatrix}^T \tag{6}$$
$$Y = \begin{bmatrix} y_1 & y_2 & \cdots & y_n \end{bmatrix}^T \tag{7}$$
$$Z = \begin{bmatrix} z_1 & z_2 & \cdots & z_n \end{bmatrix}^T \tag{8}$$

be vectors containing the $x$, $y$, and $z$ coordinates, respectively, of the deformed mesh. Extending the work in [12] to 3D, the above equation can be rewritten in matrix form as

$$\Psi_S(V) = \frac{1}{2}(K_{col}X)^T K_{col}X \\ + \frac{1}{2}(K_{col}Y)^T K_{col}Y \\ + \frac{1}{2}(K_{col}Z)^T K_{col}Z, \tag{9}$$

where $K_{col}$ is an $n_{col} \times n$ matrix, where $n_{col}$ is the number of collinear triplets. We have $n_{col} \approx 3n$, where the boundaries cause the approximation. The elements of the matrix $K_{col}$ contain the value 0, 1, or $-2$ depending upon the relationships within each triple $(i, j, k) \in E$. Each row of $K_{col}$ corresponds to a triple within the triangle mesh, and each element in the row has a value of zero except for the three locations of $i, j, k$ which contain $1, -2,$ and $1$, respectively. Each column of $K_{col}$ corresponds to a vertex within the mesh and has nonzero elements for the triplets containing the vertex.

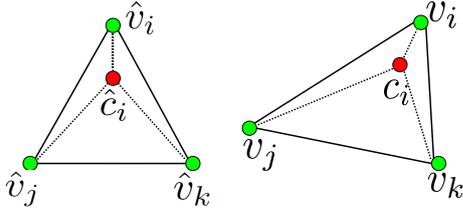

Fig. 2. Example of barycentric coordinates. The point $\hat{c}_i$ in one of the equilateral triangles in the canonical mesh (left) retains its relative position within the triangle as the triangle is deformed (right).

The above expression can be simplified as follows:

$$\Psi_S(V) = \frac{1}{2}(X^T K X + Y^T K Y + Z^T K Z), \quad (10)$$

where $K = K_{col}^T K_{col}$ is an $n \times n$ constant matrix capturing the collinear and adjacent vertices in the canonical mesh using $E$.

### B. Correspondence term

The correspondence term uses the input data of the current RGBD image to compare a possibly deformed mesh $V$ with the canonical mesh $\hat{V}$.

*1) Barycentric coordinates:* Suppose we have a point $p = (x, y, z)$ in the canonical RGBD image that happens to lie within the triangle defined by vertices $\hat{v}_i$, $\hat{v}_j$, and $\hat{v}_k$, as illustrated in Figure 2. The barycentric coordinates of $p$ are defined as the triple $(\beta_i, \beta_j, \beta_k)$ such that $p = \beta_i \hat{v}_i + \beta_j \hat{v}_j + \beta_k \hat{v}_k$ and $\beta_i + \beta_j + \beta_k = 1$. Given $p$, we compute its barycentric coordinates by solving the system of equations

$$\begin{bmatrix} \hat{x}_i - \hat{x}_k & \hat{x}_j - \hat{x}_k \\ \hat{y}_i - \hat{y}_k & \hat{y}_j - \hat{y}_k \end{bmatrix} \begin{bmatrix} \beta_i \\ \beta_j \end{bmatrix} = \begin{bmatrix} x - \hat{x}_k \\ y - \hat{x}_k \end{bmatrix} \quad (11)$$

for $\beta_i$ and $\beta_j$, then setting $\beta_k = 1 - \beta_i - \beta_j$. If $p$ lies within the triangle, then the barycentric coordinates will lie within 0 and 1, inclusive, $0 \leq \beta_i, \beta_j, \beta_k \leq 1$; and the $z$ equation will also be satisfied: $(\hat{z}_i - \hat{z}_k)\beta_i + (\hat{z}_j - \hat{z}_k)\beta_j = z - \hat{z}_k$.

Now suppose that the mesh has deformed from $\hat{V}$ to $V$. If we assume that the relative position of $p$ in the triangle remains fixed, then we can define the transformation:

$$T_V(p) = \beta_i v_i + \beta_j v_j + \beta_k v_k, \quad (12)$$

where $v_i$, $v_j$, and $v_k$ are the 3D coordinates of the deformed mesh vertices of the triangle in which $p$ lies. $T_V(p)$ yields the 3D coordinates of $p$ when the mesh is described by $V$.

*2) SURF descriptors and putative matching:* In order to capture correspondences between consecutive frames, the Speeded Up Robust Feature (SURF) detector and descriptor [41] is used. SURF is a fast and robust feature selector that is invariant to scale and rotation, providing interest locations throughout the image. SURF detects features throughout the image, but a simple foreground / background segmentation procedure using depth values is used to remove SURF features on the background, leaving only features on the triangular mesh.

Putative matching of the SURF descriptors is used to establish sparse correspondence between successive images in the video. For each pair of descriptors, the Euclidean distance is computed, and the minimum is selected as the proper feature match. Only the top fifty percent are chosen based on the minimum Euclidean distances of the feature descriptors. To improve robustness, feature matching is constrained to lie within a given threshold of a location of the feature in the previous frame of the sequence, assuming that the motion is small in consecutive frames.

*3) Correspondences:* Let $\mathcal{C} = \{(\hat{c}_i, c_i)\}_{i=1}^m$ be the set of $m$ correspondences between the canonical image and the input image. A specific correspondence $(\hat{c}_i, c_i)$ indicates that the 3D point $\hat{c}_i$ in the canonical image matches the 3D point $c_i$ in the input image, where $\hat{c}_i = (\hat{x}_{ci}, \hat{y}_{ci}, \hat{z}_{ci})$ and $c_i = (x_{ci}, y_{ci}, z_{ci})$ for $i = 1, \ldots, m$. The data term is the sum of the squared Euclidean distances between the input point coordinates and their corresponding canonical coordinates:

$$\Psi_C(V) = \frac{1}{2} \sum_{(\hat{c}_i, c_i) \in \mathcal{C}} \| c_i - T_V(\hat{c}_i) \|^2 . \quad (13)$$

### C. Depth term

The depth term measures the difference between the $z$ coordinate of each 3D vertex in the current mesh and the measured depth value. That is, it measures the distance along the ray passing through the 3D vertex to the current depth image obtained by the RGBD sensor:

$$\Psi_D(V) = \frac{1}{2} \sum_{i=1}^n |d(x_i, y_i) - z_i|^2, \quad (14)$$

where $d(x_i, y_i)$ is the value of the depth image evaluated at the position $(x_i, y_i)$, while $z_i$ is the depth of the vertex at that same position. This term is designed to ensure that the mesh fits the data even in textureless areas where no correspondences can be found.

### D. Boundary term

We define boundary vertices to be those vertices that do not have six neighbors. For every boundary vertex, we expect it to remain near the boundary of the object even as it undergoes non-rigid deformations. Figure 3 illustrates the need for the boundary term in textureless areas. To ensure this result, the boundary term captures the distances between boundary vertices and the nearest 3D boundary point within the image, as determined by the foreground / background

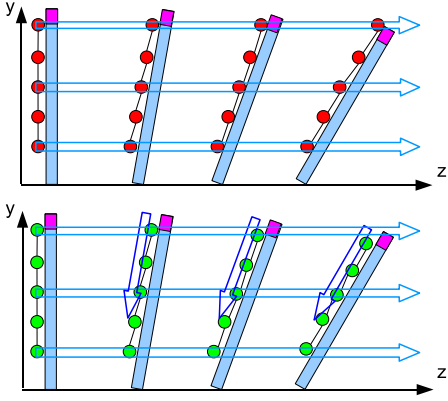

Fig. 3. An example illustrating the need for the boundary term. Top: As a vertical object is increasingly slanted over time (from left to right), the vertices deviate from their true location, due to the limitation of the depth term to induce only forces parallel to the viewing direction (indicated by horizontal light blue arrows). Bottom: With the boundary term imposing lateral forces, the mesh vertices remain in their proper locations.

segmentation procedure:

$$\Psi_B(V) = \frac{1}{2} \sum_{v_i \in \mathcal{B}} (g_d(v_i) - g_{\hat{d}}(\hat{v}_i))^2, \quad (15)$$

where $g_{\hat{d}}(\hat{v}_i)$ is the distance from the canonical vertex $\hat{v}_i$ to the nearest boundary point in the canonical image $\hat{d}$, $g_d(v_i)$ is the distance from the current vertex $v_i$ to the nearest boundary point in the current image $d$, and $\mathcal{B}$ is the set of boundary vertices. This term introduces lateral forces on the vertices to ensure the mesh fits the data in textureless regions when the object's motion has components parallel to the image plane.

### E. Energy minimization

Our goal is to locate the mesh that best explains the data while adhering to the smoothness prior. To find the configuration of minimum energy, we compute the partial derivative of the energy with respect to the vectors $X$, $Y$, and $Z$, and set the result to zero:

$$\begin{aligned}
\frac{\partial \Psi(V)}{\partial X} &= \frac{\partial \Psi_S(V)}{\partial X} + \lambda_C \frac{\partial \Psi_C(V)}{\partial X} \\
&\quad + \lambda_D \frac{\partial \Psi_D(V)}{\partial X} + \lambda_B \frac{\partial \Psi_B(V)}{\partial X} \quad (16) \\
\frac{\partial \Psi(V)}{\partial Y} &= \frac{\partial \Psi_S(V)}{\partial Y} + \lambda_C \frac{\partial \Psi_C(V)}{\partial Y} \\
&\quad + \lambda_D \frac{\partial \Psi_D(V)}{\partial Y} + \lambda_B \frac{\partial \Psi_B(V)}{\partial Y} \quad (17) \\
\frac{\partial \Psi(V)}{\partial Z} &= \frac{\partial \Psi_S(V)}{\partial Z} + \lambda_C \frac{\partial \Psi_C(V)}{\partial Z} \\
&\quad + \lambda_D \frac{\partial \Psi_D(V)}{\partial Z} + \lambda_B \frac{\partial \Psi_B(V)}{\partial Z} \quad (18)
\end{aligned}$$

Rewriting the transformation using an $n \times 1$ vector $B$ whose elements are $\beta_i$, $\beta_j$, and $\beta_k$ in the appropriate slots but zeros everywhere else:

$$T_V(p) = VB, \quad (19)$$

the partial derivatives for the correspondence term are as follows:

$$\frac{\partial \Psi_C(V)}{\partial X} = - \sum_{(\hat{c}_i, c_i) \in \mathcal{C}} (x_{ci} - X^T B) B \quad (20)$$

$$\frac{\partial \Psi_C(V)}{\partial Y} = - \sum_{(\hat{c}_i, c_i) \in \mathcal{C}} (y_{ci} - Y^T B) B \quad (21)$$

$$\frac{\partial \Psi_C(V)}{\partial Z} = - \sum_{(\hat{c}_i, c_i) \in \mathcal{C}} (z_{ci} - Z^T B) B \quad (22)$$

The partial derivatives for smoothness for straightforward:

$$\frac{\partial \Psi_S(V)}{\partial X} = KX \quad (23)$$

$$\frac{\partial \Psi_S(V)}{\partial Y} = KY \quad (24)$$

$$\frac{\partial \Psi_S(V)}{\partial Z} = KZ. \quad (25)$$

The partial derivative of the depth term is similar to that of the data term. Rewriting $z_i$ using an $n \times 1$ vector $F_i$ containing a one in the $i$th slot but zeros everywhere else:

$$z_i = Z^T F_i, \quad (26)$$

yields

$$\frac{\partial \Psi_D(V)}{\partial Z} = \sum_{i=1}^{n} (d(x_i, y_i) - Z^T F_i) F_i, \quad (27)$$

Similarly, the partial derivatives of the boundary term require rewriting $v_i$ using the same vector $F_i$:

$$g_d(v_i) = V^T F_i, \quad (28)$$

leading to

$$\frac{\partial \Psi_B(V)}{\partial X} = - \sum_{v_i \in \mathcal{B}} (\Delta X_{\hat{d}}(\hat{v}_i) - X^T F_i) F_i \quad (29)$$

$$\frac{\partial \Psi_B(V)}{\partial Y} = - \sum_{v_i \in \mathcal{B}} (\Delta Y_{\hat{d}}(\hat{v}_i) - Y^T F_i) F_i \quad (30)$$

$$\frac{\partial \Psi_B(V)}{\partial Z} = - \sum_{v_i \in \mathcal{B}} (\Delta Z_{\hat{d}}(\hat{v}_i) - Z^T F_i) F_i, \quad (31)$$

where $\hat{g}_{\hat{d}}(\hat{v}_i) = [\Delta X_{\hat{d}}(\hat{v}_i) \quad \Delta Y_{\hat{d}}(\hat{v}_i) \quad \Delta Z_{\hat{d}}(\hat{v}_i)]^T$.

In order to calculate the minimal solution, we employ the semi-implicit scheme used by Kass et al. [42]. The smoothness term is treated implicitly, while the data terms are treated explicitly. Letting $V_t$ represent the mesh at iteration $t$ and $V_{t-1}$ represent the mesh at the previous iteration $t-1$,

we have

$$\alpha(X_t - X_{t-1}) + KX_t \qquad (32)$$
$$- \lambda_C \sum_{(\hat{c}_i,c_i)\in\mathcal{C}} (x_{ci} - X_{t-1}^T B) B$$
$$- \lambda_B \sum_{v_i\in\mathcal{B}} (\Delta X_{\hat{d}}(\hat{v}_i) - X^T F_i) F_i = 0$$

$$\alpha(Y_t - Y_{t-1}) + KY_t \qquad (33)$$
$$- \lambda_C \sum_{(\hat{c}_i,c_i)\in\mathcal{C}} (y_{ci} - Y_{t-1}^T B) B$$
$$- \lambda_B \sum_{v_i\in\mathcal{B}} (\Delta Y_{\hat{d}}(\hat{v}_i) - Y^T F_i) F_i = 0$$

$$\alpha(Z_t - Z_{t-1}) + KZ_t \qquad (34)$$
$$- \lambda_C \sum_{(\hat{c}_i,c_i)\in\mathcal{C}} (z_{ci} - Z_{t-1}^T B) B$$
$$- \lambda_D \sum_{i=1}^{n} (d(x_i,y_i) - Z^T F_i) F_i$$
$$- \lambda_B \sum_{v_i\in\mathcal{B}} (\Delta Z_{\hat{d}}(\hat{v}_i) - Z^T F_i) F_i = 0,$$

where $\alpha > 0$ is the adaptation rate.

Rearranging terms leads to

$$(K + \alpha I)X_t = \alpha X_{t-1} \qquad (35)$$
$$+ \lambda_C \sum_{(\hat{c}_i,c_i)\in\mathcal{C}} (x_{ci} - X_{t-1}^T B) B$$
$$+ \lambda_B \sum_{v_i\in\mathcal{B}} (\Delta X_{\hat{d}}(\hat{v}_i) - X^T F_i) F_i$$

$$(K + \alpha I)Y_t = \alpha Y_{t-1} \qquad (36)$$
$$+ \sum_{(\hat{c}_i,c_i)\in\mathcal{C}} (y_{ci} - Y_{t-1}^T B) B$$
$$+ \lambda_B \sum_{v_i\in\mathcal{B}} (\Delta Y_{\hat{d}}(\hat{v}_i) - Y^T F_i) F_i$$

$$(K + \alpha I)Z_t = \alpha Z_{t-1} \qquad (37)$$
$$+ \sum_{(\hat{c}_i,c_i)\in\mathcal{C}} (z_{ci} - Z_{t-1}^T B) B$$
$$+ \lambda_D \sum_{i=1}^{n} (d(x_i,y_i) - Z^T F_i) F_i$$
$$+ \lambda_B \sum_{v_i\in\mathcal{B}} (\Delta Z_{\hat{d}}(\hat{v}_i) - Z^T F_i) F_i$$

Solving this equation for the unknowns $X_t$, $Y_t$, and $Z_t$ yields the desired result as an iterative sparse linear system.

## IV. EXPERIMENTAL RESULTS

We captured RGBD video sequences of shirts and posters to test our proposed method's ability to handle different non-rigid objects in a variety of scenarios. We also verified the contributions made by the novel depth and boundary energy terms to the accuracy of the estimated object configuration. For our experiments, the weights were set according to $\lambda_C$ = 1.3, $\lambda_B$ = 0.8, and $\lambda_D$ = 0.6. Further results of the system can be seen in the online video.[1]

[1] http://www.ces.clemson.edu/~stb/research/laundry_pose_estimation

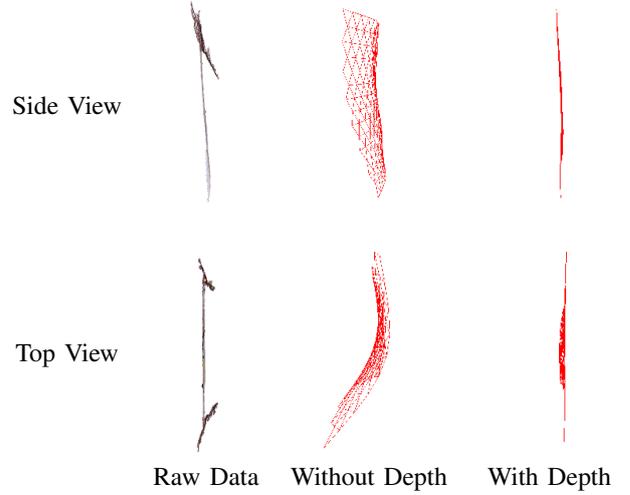

Fig. 4. The depth term improves results significantly. From left to right: the input point cloud, the estimated mesh without using the depth term, and the estimated mesh using the depth term.

### A. Illustrating the contribution of the depth term

Figure 4 compares the algorithm with and without the depth term, but with the smoothness and correspondence term in both cases. The depth term causes the triangular mesh model to adhere to the object along the $z$-axis perpendicular to the image plane. As shown in the figure, the resulting 3D meshes more accurately capture the current configuration of the object when the depth term is included. The improvement is especially noticeable in untextured areas.

### B. Illustrating the contribution of the boundary term

Figure 5 compares the algorithm with and without the boundary term, using the smoothness, correspondence, and depth term in both cases. The boundary term introduces lateral forces to cause the mesh to adhere to the object in areas near the contour of the object. The improvement resulting from the term is clear from the figure. Note also the subtle error along the top edge of the object, where the depth term alone raises the top corners (making it appear as though the top middle was sagging), as described in Figure 3.

### C. Partial self-occlusion

Figure 6 demonstrates a sequence of a T-shirt that partially occludes itself. In this experiment, it is shown that losing part of the texture located on the object does not affect the end result. Rather, the algorithm is still able to estimate the configuration of the shirt throughout the entire sequence, even without all the feature correspondences. The shirt was occluded by holding it by the shoulders and laying the lower one-third of the shirt over a chair while lowering the other two-thirds behind the rest of the chair. The figure shows the triangular mesh overlaid on the 2D RGB image, along with a side view of the 3D mesh.

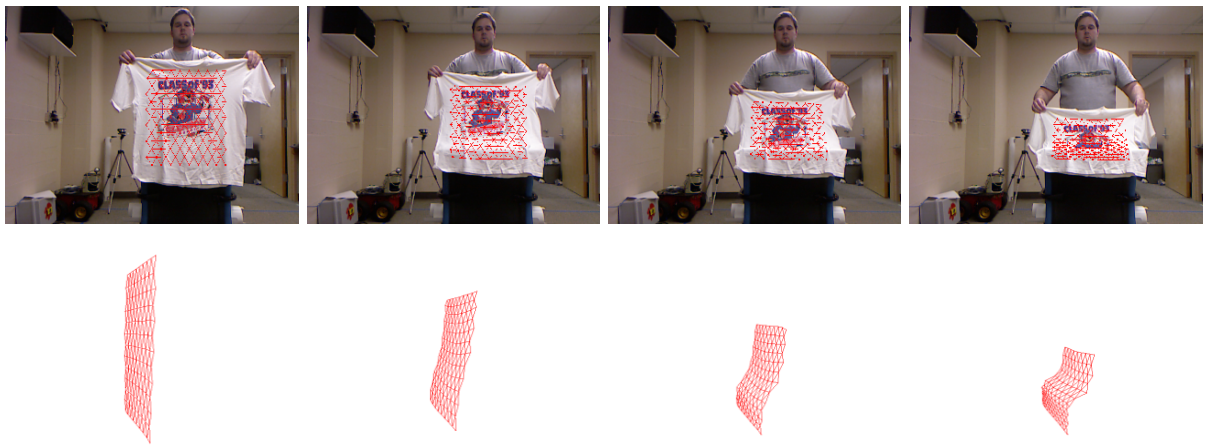

Fig. 6. Four frames of a sequence in which a shirt partially occludes itself. Top: The estimated mesh $V$ overlaid on the RGB color image. Bottom: a side of the 3D mesh with a $45^o$ pan angle.

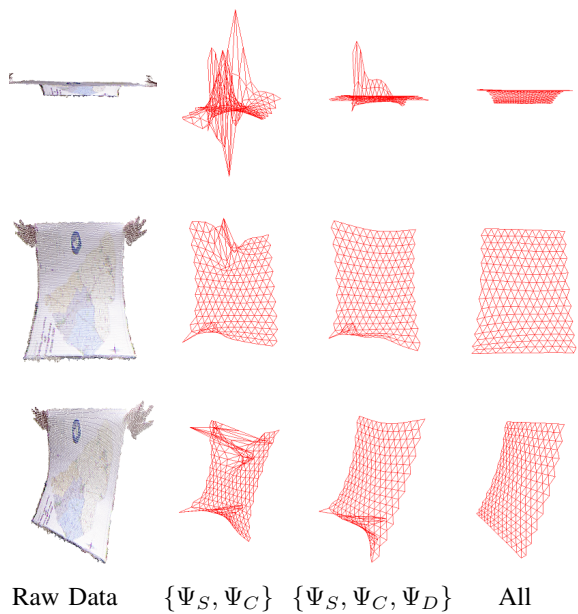

Raw Data    $\{\Psi_S, \Psi_C\}$    $\{\Psi_S, \Psi_C, \Psi_D\}$    All

Fig. 5. The improvement resulting from the various terms. From left to right: texture mapped point cloud of RGBD input data, the mesh resulting from only the smoothness and correspondence terms, the mesh from all but the boundary term, and the mesh resulting from all the terms. From top to bottom: top view, front view, and side view with $45^o$ pan.

### D. Textureless shirt sequence

Figure 7 shows a textureless T-shirt being rotated out of plane. This experiment illustrates the ability of our algorithm to estimate the configuration of a completely textureless non-rigid object without any fiducial markers. Since no SURF features were detected on the object, the energy function was required to operate without any benefit from the correspondence term. Nevertheless, the figure demonstrates that the remaining terms are able to provide an accurate estimate of the current configuration.

## V. CONCLUSION

We have presented an algorithm to estimate the 3D configuration of a nonrigid object through a video sequence using feature point correspondence, depth, and boundary information. We incorporate these terms into an energy functional that is minimized using a semi-implicit scheme, leading to an iterative sparse linear system. Results show the ability of the technique to track and estimate the configuration of nonrigid objects in a variety of scenarios. We also examine the reasons behind the need for the novel depth and boundary energy terms. In the future we plan to extend this research to handle a two-sided 3D triangular mesh that covers both the front and back of the object. We also will include color information along with depth information to provide a more accurate segmentation of the object from the background. Another step is to integrate this algorithm into a robotic system that can grasp and handle non-rigid objects in an unstructured environment.

## VI. ACKNOWLEDGEMENTS

This research was supported by the U.S. National Science Foundation under grants IIS-1017007, and IIS-0904116.

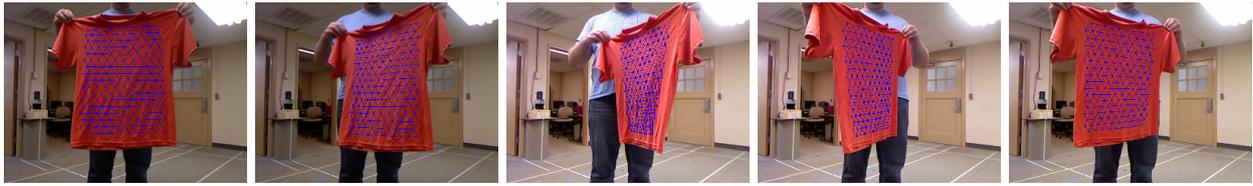

Fig. 7. Five frames from a sequence in which a completely textureless shirt with no pattern is rotated out of the image plane, while the algorithm estimates the configuration of the shirt as a triangular mesh estimate. Even without data correspondences, the proposed energy function is able to accurately estimate the configuration of a dynamic non-rigid object.